\def\year{2020}\relax
\documentclass[letterpaper]{article} 
\usepackage{aaai20}
\usepackage{times}
\usepackage{helvet}
\usepackage{courier}
\usepackage{url}
\usepackage{graphicx}
\usepackage{breqn}
\usepackage{epstopdf}
\title{FedVision: An Online Visual Object Detection Platform Powered by \\Federated Learning}
\author{
Yang Liu\textsuperscript{\rm 1*},
Anbu Huang\textsuperscript{\rm 1},
Yun Luo\textsuperscript{\rm 2,3*},
He Huang\textsuperscript{\rm 3},
Youzhi Liu\textsuperscript{\rm 1},
Yuanyuan Chen\textsuperscript{\rm 4,5},\\
\Large\textbf{Lican Feng\textsuperscript{\rm 3},}
\textbf{Tianjian Chen\textsuperscript{\rm 1},}
\textbf{Han Yu\textsuperscript{\rm 4,5*}} and
\textbf{Qiang Yang\textsuperscript{\rm 1,2}}
\\ 
\textsuperscript{\rm 1}Department of AI, WeBank, Shenzhen, China\\
\textsuperscript{\rm 2}Department of Computer Science and Engineering, Hong Kong University of Science and Technology, Hong Kong\\
\textsuperscript{\rm 3}Extreme Vision Ltd, Shenzhen, China\\
\textsuperscript{\rm 4}School of Computer Science and Engineering, Nanyang Technological University, Singapore\\
\textsuperscript{\rm 5}Joint NTU-WeBank Research Centre on FinTech, NTU, Singapore\\
*Corresponding Authors: yangliu@webank.com, lauren.luo@extremevision.mo, han.yu@ntu.edu.sg
}
 
\begin{document}

\maketitle

\begin{abstract}
Visual object detection is a computer vision-based artificial intelligence (AI) technique which has many practical applications (e.g., fire hazard monitoring). However, due to privacy concerns and the high cost of transmitting video data, it is highly challenging to build object detection models on centrally stored large training datasets following the current approach. Federated learning (FL) is a promising approach to resolve this challenge. 
Nevertheless, there currently lacks an easy to use tool to enable computer vision application developers who are not experts in federated learning to conveniently leverage this technology and apply it in their systems. In this paper, we report \textit{FedVision} - a machine learning engineering platform to support the development of federated learning powered computer vision applications. The platform has been deployed through a collaboration between \textit{WeBank} and \textit{Extreme Vision} to help customers develop computer vision-based safety monitoring solutions in smart city applications. Over four months of usage, it has achieved significant efficiency improvement and cost reduction while removing the need to transmit sensitive data for three major corporate customers. To the best of our knowledge, this is the first real application of FL in computer vision-based tasks.
\end{abstract}

\section{Introduction}
Object detection is one of the most important applications of computer vision in the field of artificial intelligence (AI). It has been widely adopted by practical applications such as safety monitoring.
Over the past decade, visual object detection technologies have experienced significant advancement as deep learning techniques developed \cite{Ren:2017:FRT:3101720.3101780,DBLP:journals/corr/abs-1804-02767,DBLP:journals/corr/abs-1807-05511}. The prevailing training approach requires centralized storage of training data in order to obtain powerful object detection models. The typical workflow of training an object detection algorithm this way is shows in Figure \ref{fig-trad-workflow}. Under such an approach, each data owner (i.e. user) annotates visual data from cameras and uploads these labelled training data to a central database (e.g., a cloud server) for model training. Once the model has been trained, it can be used to perform inference tasks. In this way, the users have no control over how the data would be used once they are transmitted to the central database. Besides, centralized model training has the following limitations:

\begin{figure}[!t]
	\centering
	\includegraphics[width=1\columnwidth]{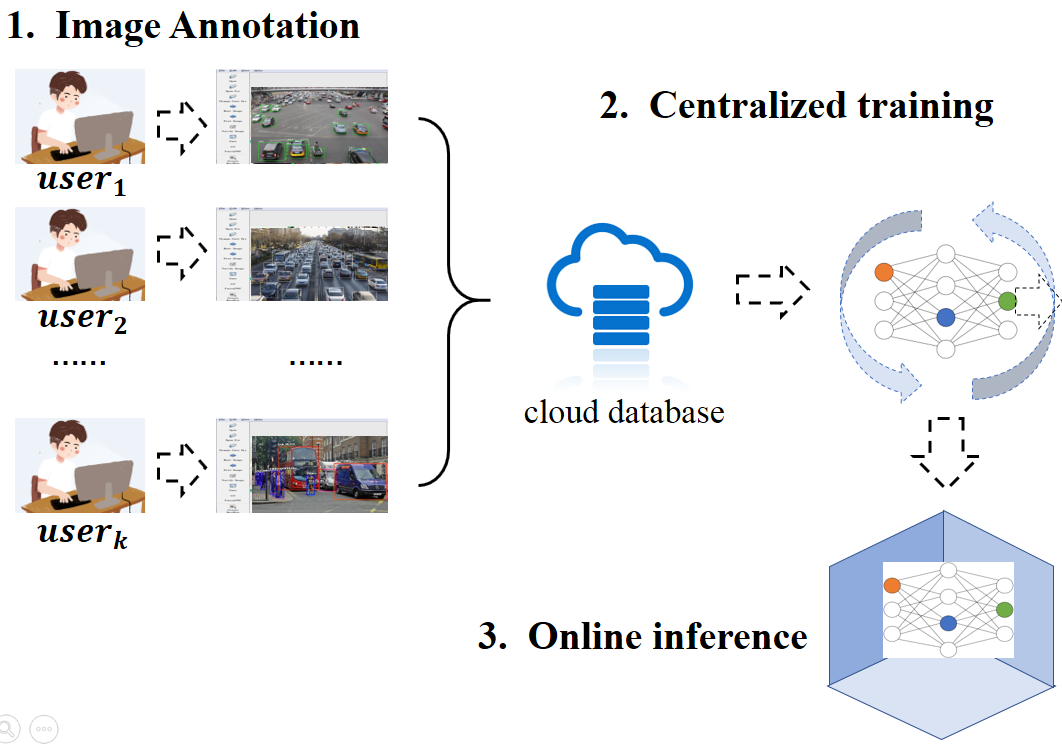}
	\caption{A typical workflow for centralized training of an object detector.}
	\label{fig-trad-workflow}
\end{figure}

\begin{enumerate}
\item It is difficult to share data across organizations due to liability concerns. Increasingly strict data sharing regulations (e.g., the General Data Protection Regulation (GDPR) \cite{Voigt:2017:EGD:3152676}) restrict data sharing across organizations;
\item The whole process takes a long time and depends on when the next round of off-line training occurs. When users accumulate new data, they must upload the data to the central training server and wait for the next round of training to happen, which they cannot control, in order to receive an updated model. This also leads to the problem of lagging feedbacks which delay the correction of any errors in model inference ; and
\item The amount of data required to train a useful object detector tends to be large, and uploading them to a central database incurs significant communication cost.
\end{enumerate}

Due to these limitations, in commercial settings, the following anecdotal conversation between a customer (C) and an AI solution provider (P) can often be heard:
\begin{itemize}
\item [C]: \textit{``We need a solution for detecting flames in a factory floor from surveillance videos to provide early warning of fire.''}
\item [P]: \textit{``No problem. We will need some data from you to train a flame detection model.''}
\item [C]: \textit{``Of course. We have a lot of data, both images and videos with annotations.''}
\item [P]: \textit{``Great! Please upload your datasets to our server.''}
\item [C]: \textit{``Such data contain sensitive information, we can't pass them on to a third party.''}
\item [P]: \textit{``We could send our engineers to work with your dataset on-site, but this will incur additional costs.''}
\item [C]: \textit{``Well, this looks expensive and is beyond our current budget ... ...''}
\end{itemize}

This challenging situation urges the AI research community to seek new methods of training machine learning models. Federated Learning (FL) \cite{DBLP:journals/tist/YangLCT19}, which was first proposed by Google in 2016 \cite{45648}, is a promising approach to resolve this challenge. The main idea is to build machine learning models based on distributed datasets while keeping data locally stored, hence preventing data leakage and minimizing communication overhead. FL balances performance and efficiency issues while preventing sensitive data from being disclosed. In essence, FL is a collaborative computing framework. FL models are trained via model aggregation rather than data aggregation. Under the federated learning framework, we only need to train a visual object detection model locally at a data owner's site, and upload the model parameters to a central server for aggregation, without the need to upload the actual training dataset.

However, there currently lacks an easy to use tool to enable developers who are not experts in federated learning to conveniently leverage this technology in practical computer vision applications. In order to bridge this gap, we report \textit{FedVision} - a machine learning engineering platform to support easy development of federated learning powered computer vision applications. It currently supports a proprietary federated visual object detection algorithm framework based on YOLOv3 \cite{DBLP:journals/corr/abs-1804-02767}, and allows end-to-end joint training of object detection models with locally stored datasets from multiple clients. The user interaction for learning task creation follows a simplified design which does not require users to be familiar with the FL technology in order to make use of it.

The platform has been deployed through a collaboration between \textit{WeBank}\footnote{\url{https://www.webank.com/en/}} and \textit{Extreme Vision}\footnote{\url{https://www.extremevision.com.cn/?lang=en_US}} since May 2019. It has been used by three large-scale corporate customers to develop computer vision-based safety monitoring solutions in smart city applications. After four months of usage at the time of submission of this paper, the platform has helped the customers significantly improve their operational efficiency and reduce their costs, while eliminating the need to transmit sensitive data around. To the best of our knowledge, this is the first industry application of federated learning in computer vision-based tasks.

\section{Application Description}
\begin{figure}[ht]
	\centering
	\includegraphics[width=1\columnwidth]{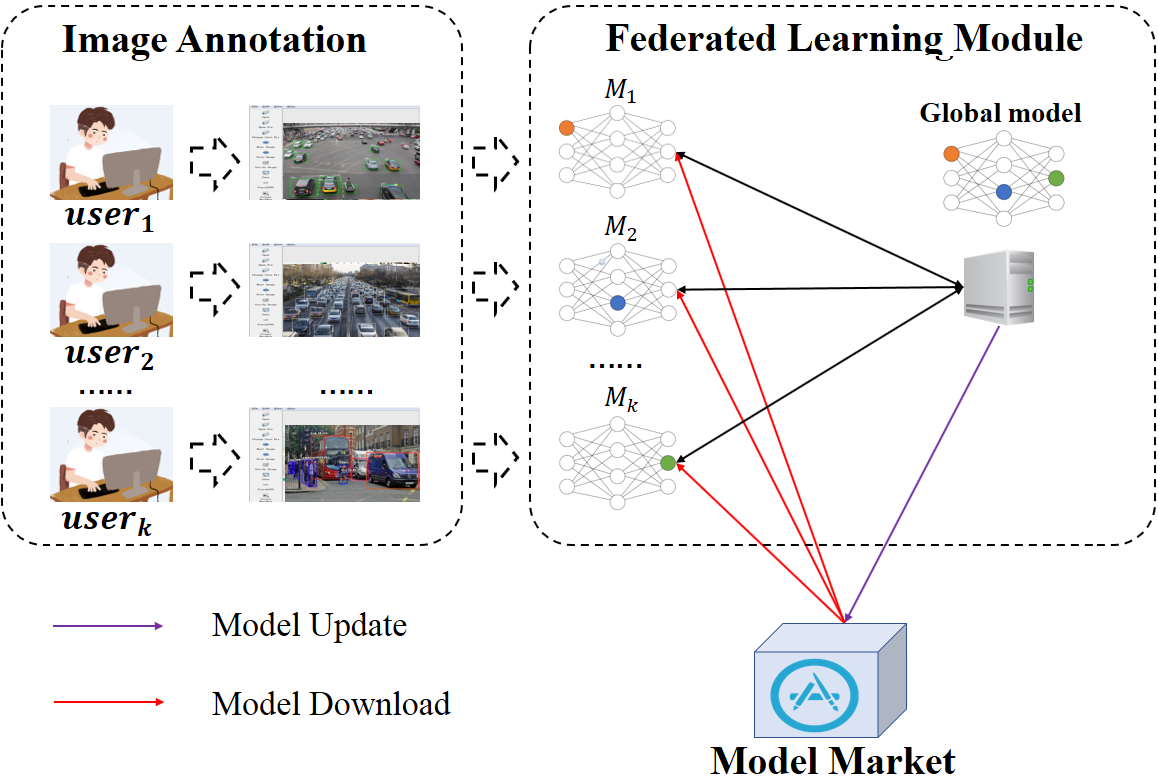}
	\caption{The FedVision workflow for training a visual object detection algorithm with data from multiple users.}
	\label{fig-workflow}
\end{figure}

In this section, we describe the system design of FedVision. The new workflow for training a visual object detection algorithm under FedVision is as shows in Figure \ref{fig-workflow}. It consists of three main steps: 1) crowdsourced image annotation, 2) federated model training; and 3) federated model update.

\subsection{Crowdsourced Image Annotation}
\begin{figure*}[ht]
	\centering
	\includegraphics[width=1\linewidth]{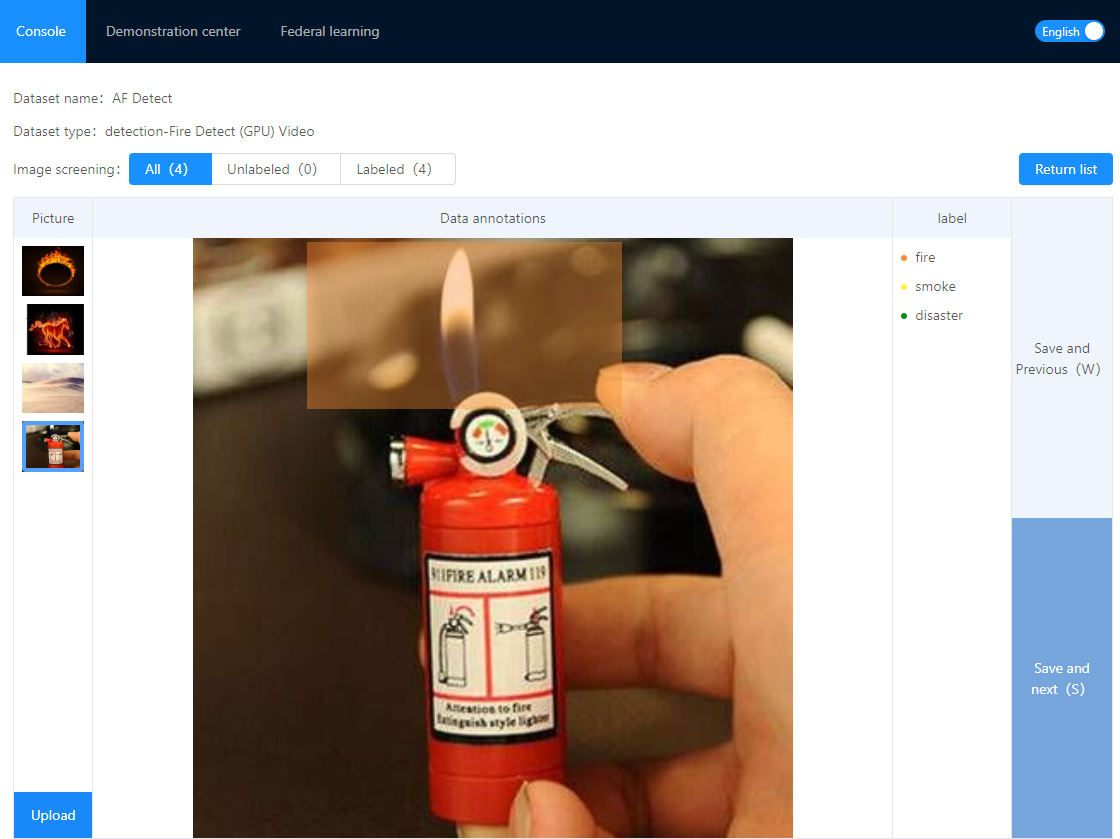}
	\caption{The image annotation module of FedVision.}
	\label{fig-label}
\end{figure*}

This module is designed for data owners to easily label their locally stored image data for FL model training.
A user can annotate a given image on his local device by using the interface provided by FedVision (Figure \ref{fig-label}) to easily specify each bounding box and the corresponding label information. FedVision adopts the Darknet\footnote{\url{https://pjreddie.com/darknet/}} model format for annotation. Thus, each row represents information for a bounding box in the following form:
\centerline{\{label   $x$   $y$   $w$   $h$\}}
where ``label'' denotes the category of objects, $(x, y)$ represents the center of the bounding box, $(w, h)$ represents the width and height of the bounding box.

The process only requires the user to be able to visually identify where the objects of interest (e.g., flames) are in a given image, use the mouse to draw the bounding box, and assign it to a category (e.g., fire, smoke, disaster). Users do not need to possess knowledge about federated learning. The annotation file is automatically mapped to the appropriate system directory for model training by FedVision. With this tool, the task of labelling training data can be easily distributed among data owners in a way similar to crowdsourcing \cite{Doan-et-al:2011}, and thereby, reducing the burden of image annotation on the entity coordinating the learning process. It can also be used to support online learning as new image data arrive sequentially over time from the cameras.

\subsection{Horizontal Federated Learning (HFL)}
\begin{figure}[!b]
	\centering
	\includegraphics[width=1\columnwidth]{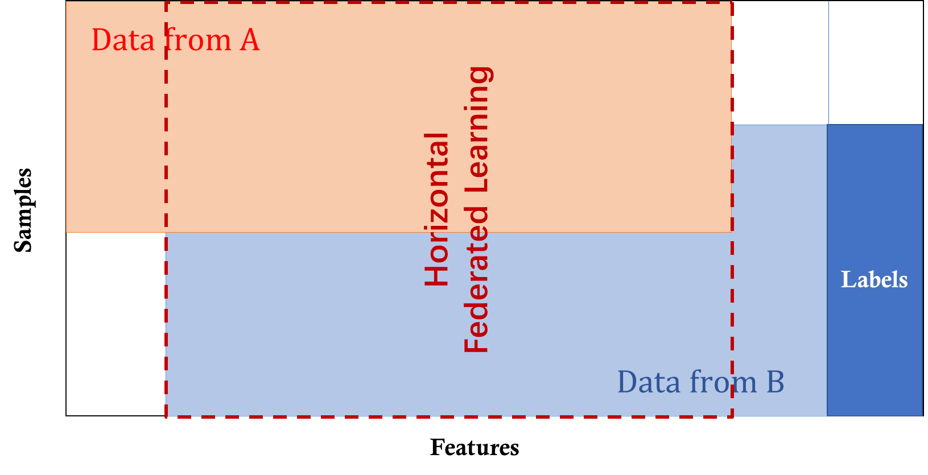}
	\caption{The concept of horizontal federated learning (HFL)\cite{DBLP:journals/tist/YangLCT19}}
	\label{fig:hfl}
\end{figure}

In order to understand the federated learning technologies incorporated into the FedVision platform, we first introduce the concept of horizontal federated learning (HFL). HFL, also known as sample-based federated learning, can be applied in scenarios in which datasets share the same feature space, but differ in sample space (Figure \ref{fig:hfl}). In other words, different parties own datasets which are of the same format but collected from different sources.

HFL is suitable for the application scenario of FedVision since it aims to help multiple parties (i.e. data owners) with data from the same feature space (i.e. labelled image data) to jointly train federated object detection models. The word ``horizontal'' comes from  the term ``horizontal partition'', which is widely used in the context of the traditional tabular view of a database (i.e. rows of a table are horizontally partitioned into different groups and each row contains the complete set of data features). We summarize the conditions for HFL for two parties, without loss of generality, as follows.
\begin{equation}
\mathcal{X}_{a}=\mathcal{X}_{b},\ \ \mathcal{Y}_{a}=\mathcal{Y}_{b}, \ \  I_{a}\neq I_{b},\ \ \forall \mathcal{D}_{a}, \mathcal{D}_{b}, a\neq b
\end{equation}
where the data features and labels of the two parties, $(\mathcal{X}_{a}, \mathcal{Y}_{a})$ and $(\mathcal{X}_{b}, \mathcal{Y}_{b})$, are the same, but the data entity identifiers $I_{a}$ and $I_{b}$ can be different. $\mathcal{D}_{a}$ and $\mathcal{D}_{b}$ denote the datasets owned by Party $a$ and Party $b$, respectively.

Under HFL, data collected and stored by each party are no longer required to be uploaded to a common server to facilitate model training. Instead, the model framework is sent from the federated learning server to each party, which then uses the locally stored data to train this model. After training converges, the encrypted model parameters from each party are sent back to the server. They are then aggregated into a global federated model. This global model will eventually be distributed to the parties in the federation to be used for inference in subsequent operations.

\subsection{Federated Model Training}
\begin{figure}[!b]
	\centering
	\includegraphics[width=1\columnwidth]{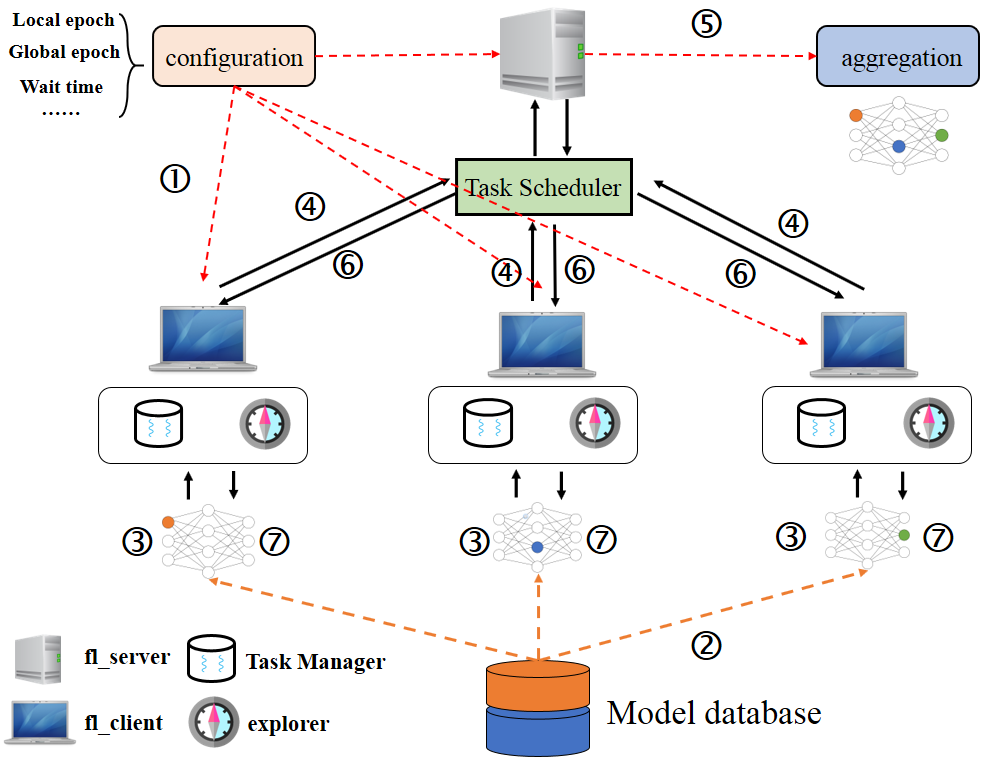}
	\caption{The system architecture of the federated model training module.}
	\label{fig-fl_fr}
\end{figure}

The FedVision platform includes an AI Engine which consists of the federated model training and the federated model update modules.
From a system architecture perspective, the federated model training module consists of the following six components (Figure \ref{fig-fl_fr}):

\begin{enumerate}
    \item Configuration: it allows users to configure training information, such as the number of iterations, the number of reconnections, the server URL for uploading model parameters and other key parameters.
    \item Task Scheduler: it performs global dispatch scheduling which is used to coordinate communications between the federated learning server and the clients in order to balance the utilization of local computational resources during the federated model training process. The load-balancing approach is based on \cite{Yu-et-al:2017SciRep} which jointly considers clients' local model quality and the current load on their local computational resources in an effort to maximize the quality of the resulting federated model.
    \item Task Manager: when multiple model algorithms are being trained concurrently by the clients, this component coordinates the concurrent federated model training processes.
    \item Explorer: this component monitors the resource utilization situation on the client side (e.g., CPU usage, memory usage, network load, etc.), so as to inform the Task Scheduler on its load-balancing decisions.
    \item FL\_SERVER: this is the server for federated learning. It is responsible for model parameter uploading, model aggregation, and model dispatch which are essential steps involved in federated learning \cite{DBLP:journals/corr/abs-1902-01046}.
    \item FL\_CLIENT: it hosts the Task Manager and Explorer components and performs local model training which is also an essential step involved in federated learning \cite{DBLP:journals/corr/abs-1902-01046}.
\end{enumerate}

\subsection{Federated Model Update}
\begin{figure}[ht]
	\centering
	\includegraphics[width=1\columnwidth]{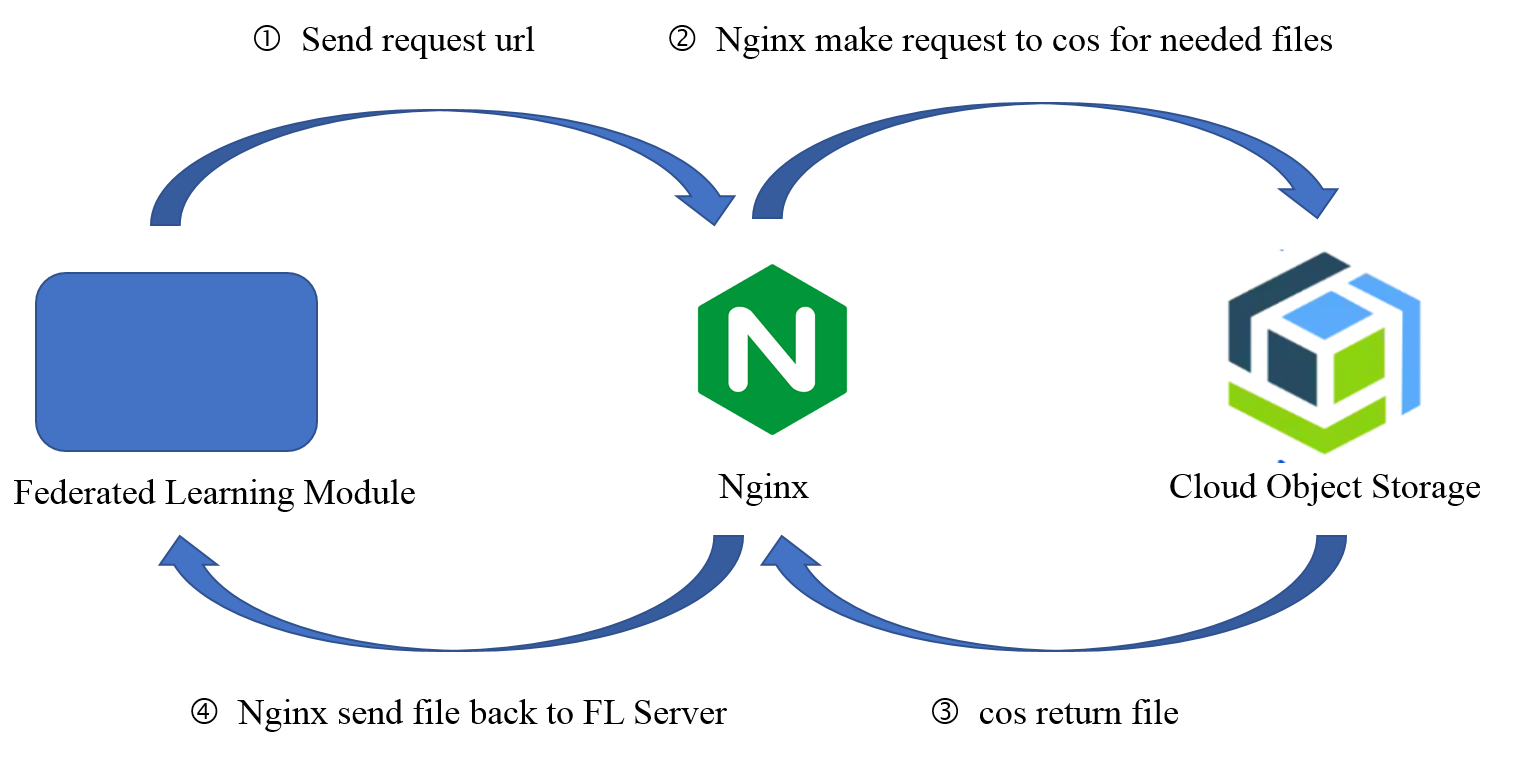}
	\caption{Cloud Object Storage (COS).}
	\label{fig-COS}
\end{figure}

After local model training, the model parameters from each user's FL\_CLIENT are transmitted to the FL\_SERVER. The updated model parameters need to be stored. The number of such model parameter files, and thus the storage size required, increases with the rounds of training operations. FedVision adopts Cloud Object Storage (COS) to store practically limitless amounts of data easily and at an affordable cost. The workflow of storing federated object detection model parameters via COS is shown in Figure \ref{fig-COS}.

FedVision provides a model aggregation approach to combine local model parameters into a federated object detection model. Details of the federated model training and federated model update components of the FedVision AI Engine are provided in the next section.

\section{Uses of AI Technology}
In this section, we discuss the AI Engine of FedVision. We explain the federated object detection model which is at the core of the FedVision AI Engine, and the neural network compression technique adopted by FedVision to optimize the efficiency of transmitting federated model parameters.

\subsection{Federated Object Detection Model Training}
From the Regions with Convolutional Neural Network features (R-CNN) model \cite{RCNN2014} to the latest YOLOv3 model \cite{DBLP:journals/corr/abs-1804-02767}, deep learning-based visual object detection approaches have experience significant improvement in terms of accuracy and efficiency. From the perspective of model workflow, these approaches can be divided into two major categories: 1) one-stage approaches, and 2) two-stage approaches.

In a typical two-stage approach, the algorithm first generates candidate regions of interest, and then extracts features using CNN to perform classification of regions while improving positioning. In a typical one-stage approach, no candidate region needs to be generated. Instead, the algorithm treats the problems of bounding box positioning and classification of the regions as regression tasks. In general, two-stage approaches produce more accurate object detection results, while one-stage approaches are more efficient. As the application scenarios for FedVision prioritize efficiency over accuracy, we adopt YOLOv3, which is a one-stage approach, as the basic object detection model and implement a federated learning version of it -- \textit{Federated YOLOv3 (FedYOLOv3)} -- in our platform. With one round of end-to-end training, FedYOLOv3 can identify the position of the bounding box as well as the class for the target object in an image.

\begin{figure}[!t]
	\centering
	\includegraphics[width=0.5\columnwidth]{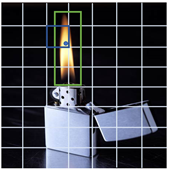}
	\caption{Flame detection with YOLOv3.}
	\label{fig-flame}
\end{figure}

The approach of YOLOv3 can be summarized as follows. Given an image, such as the image of a flame as shown in Figure \ref{fig-flame}, it is first divided into an $S\times S$ grid with each grid being used for detecting the target object with its centre located in the given grid (the blue square grid in Figure \ref{fig-flame} is used to detect flames). For each grid, the algorithm performs the following computations:
\begin{enumerate}
    \item Predicting the positions of $B$ bounding boxes. Each bounding box is denoted by a 4-tuple $\langle x,y,w,h \rangle$, where $(x,y)$ is the coordinate of the centre, and $(w,h)$ represent the width and height of the bounding box, respectively.
    \item Estimating the confidence score for the $B$ predicted bounding boxes. The confidence score consists of two parts: 1) whether a bounding box contains the target object, and 2) how precise the boundary of the box is. The first part can be denoted as $p(obj)$. If the bounding box contains the target object, then $p(obj)=1$; otherwise, $p(obj)=0$. The precision of the bounding box can be measured by its intersection-over-union (IOU) value, $IOU$, with respect to the ground truth bounding box. Thus, the confidence score can be expressed as $\theta=p(obj)\times IOU$.
    \item Computing the class conditional probability, $p(c_{i}|obj)\in[0,1]$, for each of the $C$ classes.
\end{enumerate}

The loss function of YOLOv3 consists of three parts:
\begin{enumerate}
  \item Class prediction loss, which is expressed as
    \begin{equation}
    \sum_{i=0}^{S^{2}}1_{i}^{obj}\sum_{c}(p_{i}(c)-\hat{p}_{i}(c))^{2}
    \end{equation}
    where $p_{i}(c)$ represents the probability of grid $i$ belonging to class $c$, and $\hat{p}_{i}(c)$ denotes the probability that grid $i$ is predicted to be belonging to class $c$ by the model. $1_{i}^{obj}=1$ if grid $i$ contains the target object; otherwise, $1_{i}^{obj}=0$.
  \item Bounding box coordinate prediction loss, which is expressed as
    \begin{dmath}
    \lambda_{coord}\sum_{i=0}^{S^{2}}\sum_{j=0}^{B}1_{ij}^{obj}[(x_{ij}-\hat{x}_{ij})^{2}+(y_{ij}-\hat{y}_{ij})^{2}]
    +\lambda_{coord}\sum_{i=0}^{S^{2}}\sum_{j=0}^{B}1_{ij}^{obj}[(w_{ij}-\hat{w}_{ij})^{2}+(h_{ij}-\hat{h}_{ij})^{2}]
    \end{dmath}
    where $\langle x_{ij},y_{ij},w_{ij},h_{ij} \rangle$ denote the ground truth bounding box coordinates, and $\langle \hat{x}_{ij},\hat{y}_{ij},\hat{w}_{ij},\hat{h}_{ij} \rangle$ denote the predicted bounding box coordinates.
  \item Confidence score prediction loss, which is expressed as
    \begin{equation}
    \sum_{i=0}^{S^{2}}\sum_{j=0}^{B}1_{ij}^{obj}(\theta_{i}-\hat{\theta}_{i})^{2}+\lambda_{\neg obj}\sum_{i=0}^{S^{2}}\sum_{j=0}^{B}1_{ij}^{\neg obj}(\theta_{i}-\hat{\theta}_{i})^{2}.
    \end{equation}
\end{enumerate}
Here, $\lambda_{coord}$ and $\lambda_{\neg obj}$ are well studied hyper-parameters of the model. The default value of them have been pre-configured into the platform.

Once the users use the FedVision image annotation tool to label their local training datasets, they can join the FedYOLOv3 model training process as described in the previous section. Once the local model converges, a user $a$ can initiate the transfer of the current local model parameters (in the form of the weight matrix $W_{a}(t)$) to FL\_SERVER in a secure encrypted manner through his FL\_CLIENT. The HFL module in FedVision operates in rounds. After each round of learning elapses, FL\_SERVER performs federated averaging \cite{FedAvg2016} to compute an updated global weight matrix for the model, $W(t)$:
\begin{equation}
W(t)=\frac{1}{N}\sum_{a=1}^{N}W_{a}(t).
\end{equation}
FL\_SERVER then sends the updated $W(t)$ to the $N$ participating FL\_CLIENTs so that they can enjoy the benefits of an updated object detection model trained with everyone's dataset in essence. In this way, FedYOLOv3 can rapidly respond to any potential errors in model inference.

\subsection{Model Compression}
\begin{figure}[ht]
	\centering
	\includegraphics[width=1\columnwidth]{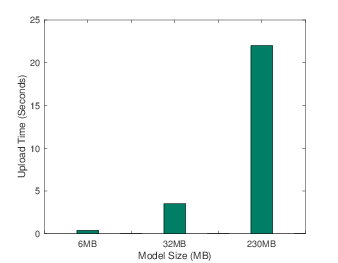}
	\caption{Time for uploading federated model parameters of different sizes.}
	\label{fig-upload}
\end{figure}

Uploading model parameters to the FL\_SERVER can be time consuming due to network bandwidth constraints. Figure \ref{fig-upload} shows the upload time required for uploading federated model parameters of different sizes. For example, when the network bandwidth is about 15MB/sec, it takes more than 20 seconds to upload a $W_{a}(t)$ of 230MB in size.

However, during the federated model training, different model parameters might have different contributions towards model performance. Thus, neural network compression can be performed to reduce the sizes of the network by pruning less useful weight values while preserving model performance \cite{DBLP:journals/corr/abs-1710-09282}. In FedVision, we apply network pruning to compress the federated model parameters and speed up transmission \cite{DBLP:conf/iclr/2016}.

Let $M^{i,k}$ be the model parameter matrix from the $i$-th user after completing the $k$-th iteration of federated model training. Let $M^{i, k}_{j}$ be the $j$-th layer of $M^{i,k}$. We denote the sum of the absolute values of all parameters in the $j$-th layer as $|\sum{M^{i,k}_{j}}|$. The contribution of the $j$-th layer to the overall model performance, $v(j)$, can be expressed as:
\begin{equation}
v(j)=\left|\sum{M^{i,k}_{j}}-\sum{M^{i,(k-1)}_{j}}\right|
\end{equation}
The larger the value of $v(j)$, the greater the impact of layer $j$ on the model. FL\_CLIENT ranks the $v(j)$ values of all layers in the model in descending order, and selects only the parameters of the first $n$ layers to be uploaded to the FL\_SERVER for federated model aggregation. A user can set the desired value for $n$ through FedVision.

\section{Application Use and Payoff}
FedVision has been deployment through a collaboration between Extreme Vision and WeBank since May 2019. It has been used by three large-scale corporate customers: 1) China Resources (CRC)\footnote{\url{https://en.crc.com.cn/}}, 2) GRG Banking\footnote{\url{http://www.grgbanking.com/en/}}, 3) State Power Investment Corporation (SPIC)\footnote{\url{http://eng.spic.com.cn/}}.

CRC has business interests in consumer products, healthcare, energy services, urban construction and operation, technology and finance. It has more than 420,000 employees. FedVision has been used to help it detect multiple types of safety hazards via cameras in more than 100 factories.

GRG Banking is a globally renowned AI solution provider in financial self-service industry in the world. It has more than 300,000 equipment (e.g., ATMs) deployed in over 80 countries. FedVision has been used to help it monitor suspicious transaction behaviours via cameras on the equipment.

SPIC is the world's largest photovoltaic power generation company which facilities in 43 countries. FedVision has been used to help it monitor the safety of more than 10,000 square meters of photovoltaic panels.

Over the four months of usage, FedVision has achieved the following business improvements:
\begin{enumerate}
    \item \textit{Efficiency}: in the flame identification system of CRC, to improve a model, at least 1,000 sample images were needed. The entire procedure generally required 5 labellers for about 2 weeks, including the time of testing and packaging. Thus, the total time for model optimization can be up to 30 days. In subsequent operations, the procedure would be repeated. With FedVision, the system administrator can finish labeling the images by himself. The time of model optimization is reduced by more than 20 days, saving labor cost.
    \item \textit{Data Privacy}: under FedVision, image data do not need to leave the machine with which they are collected to facilitate model training. In the case of GRGBanking, to 10,000 photos were required to train its model. Each photo is around 1 MB in size. The 10,000 photos used to require 2 to 3 days to be collected and downloaded to a central location. During this process, the data would go through 2 to 3 locations and are at risk of being exposed. With the help of FedVision, GRGBanking can leverage the local storage and computational resources at their ATM equipment to train a federated suspicious activity detection model, thereby reducing the risk of data exposure.
    \item \textit{Cost}: in the generator monitoring system of SPIC, a total of 100 channels of surveillance videos are in place in one generator facility. Under the data transmission rate of 512 KB/sec for synchronous algorithm analysis and optimization, these 100 channels require at least 50 MB/sec of network bandwidth if image data need to be sent. This is expensive to implement on an industry scale. With FedVision, the network bandwidth required for model update is significantly reduced to less than 1 MB/sec.
\end{enumerate}

The improvements brought about by the FedVision platform has significantly enhanced the operations of the customers and provided them with competitive business advantages.

\section{Application Development and Deployment}
\begin{figure*}[!t]
	\centering
	\includegraphics[width=1\linewidth]{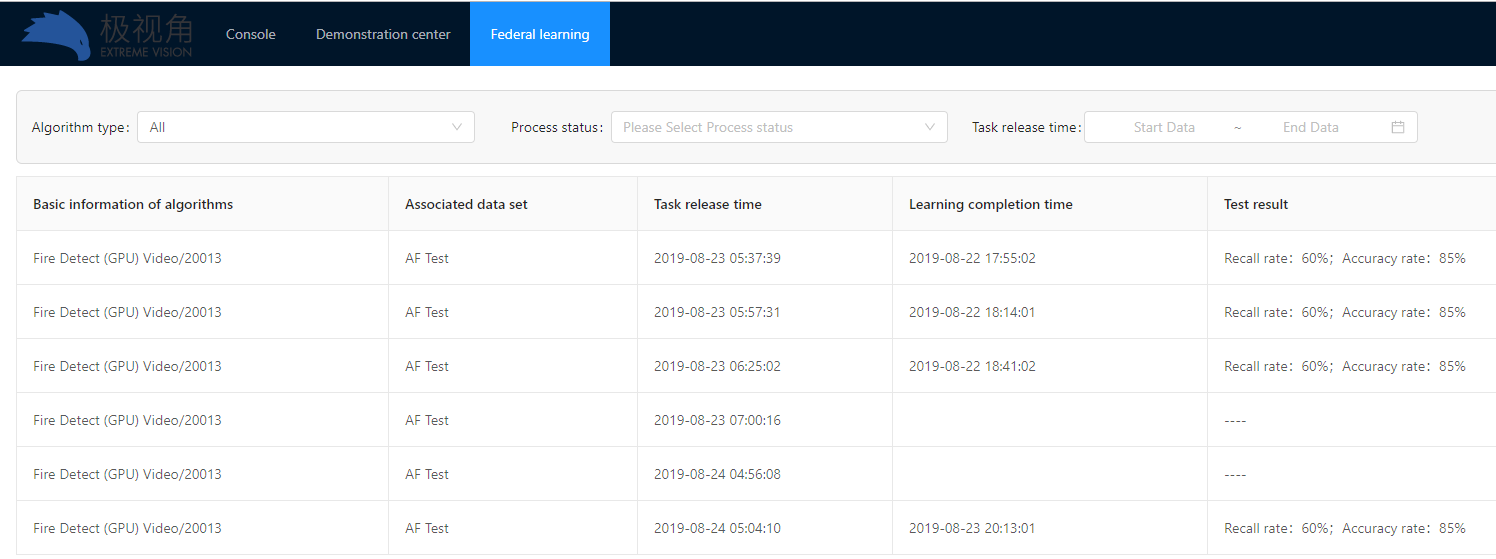}
	\caption{Monitoring multiple rounds of federated model training on FedVision.}
	\label{fig-UI}
\end{figure*}

The FedVision platform was developed using Python and C programming languages by WeBank, Shenzhen, China. When developing the AI Engine, we have evaluated multiple potential approaches which are capable of fulfilling our design objectives while disallowing the explicit sharing of locally stored camera data. These include secure multi-party computation (MPC), differential privacy (DP), and federated learning (FL). The decision for selecting FL is based on the following considerations:
\begin{enumerate}
\item In recent decades, many privacy-preserving machine learning methods have been proposed. They are mostly based on secure MPC \cite{Yao:1982:PSC:1398511.1382751,Yao:1986:GES:1382439.1382944}. The goal is to protect data privacy, and explore how to calculate a conversion function safely without a trusted third party. However, the transmission efficiency between multiple parties is very low. This makes unsuitable for our application which not only demands high efficiency, but also requires the involvement of a trusted third party (i.e. Extreme Vision Ltd) to fulfil certain business objectives.
\item Differential Privacy \cite{Dwork:2006:DP:2097282.2097284,Wang-et-al:2019} aims to protect sensitive data by adding noise into the dataset in such a way that preserves the overall distribution of the data. A trade-off needs to be made between the strength of privacy protection versus the usefulness of the resulting dataset for inference tasks. However, DP still requires data aggregation for model training. This not only violates the requirements by privacy protection laws such as GDPR, but also incurs high communication overhead as the datasets are artificially enlarged with the added noise.
\item Federated learning is the best available technology for building the AI Engine of the FedVision platform. It does not require data aggregation for model training. Thus, it not only preserves data privacy, but also significantly reduces communication overhead. In addition, with a wide range of model aggregation algorithms (e.g., federated averaging \cite{FedAvg2016}), FL provides better support for extending deep learning-based models which are widely used in visual object detection tasks.
\end{enumerate}
Therefore, FL has been selected to implement the AI Engine of FedVision.

Once the user completed the annotation of his local dataset and joined the construction of a federated object detection model on FedVision, the rest of the processes are taken care of by the platform automatically. The user can conveniently monitor the progress of different rounds of federated model training through the user interface as shown in Figure \ref{fig-UI}. As it is developed for use in China, the language used in the actual production user interface is Chinese. The version shown in Figure \ref{fig-UI} is for readers who do not speak Chinese. A video demonstration of the functionalities of the FedVision platform can be accessed online\footnote{\url{https://youtu.be/yfiO3NnSqFM}}.

\section{Maintenance}
As time goes by, there are additions of new types of computer vision-based tasks, changes in personnel access rights, and changes in operating parameters in
FedVision. Since the platform architecture follows are modular design approach around tasks and personnel to achieve separation of concern with respect to the AI Engine, such updates can be performed without affecting the AI Engine. 
Since deployment in May 2019, there has not been any AI maintenance task.

\section{Conclusions and Future Work}
In this paper, we report our experience addressing the challenges of building effective visual object detection models with image data owned by different organizations through federated learning. The deployed FedVision platform is an end-to-end machine learning engineering platform for supporting easy development of FL-powered computer vision applications. The platform has been used by three large-scale corporate customers to develop computer vision-based safety hazard warning solutions in smart city applications. Over four months of deployment, it has helped the customers improve their operational efficiency, achieve data privacy protection, and reduced cost significantly. To the best of our knowledge, this is the first industry application of federated learning in computer vision-based tasks. It has the potential to help computer vision-based applications to comply with stricter data privacy protection laws such as GDPR \cite{Voigt:2017:EGD:3152676}.

Currently, the FedVision platform allows users to easily utilize the FedYOLOv3 algorithm. In subsequent work, we will continue to incorporate more advanced FL algorithms (e.g. federated transfer learning \cite{Liu-et-al:2019,Gao-et-al:2019}) into the platform to deal with more complex learning tasks. We are also working on improving the explainability of models through visualization \cite{Wei-et-al:2019} in the platform in order to build trust with the users \cite{Yu-et-al:2018IJCAI}. We will enhance the platform with incentive mechanisms \cite{Cong-et-al:2019} to enable the emergence of a sustainable FL ecosystem over time.

\section{Acknowledgments}
This research is supported by the Nanyang Assistant Professorship (NAP); AISG-GC-2019-003; NRF-NRFI05-2019-0002; NTU-WeBank JRI (NWJ-2019-007); and the R\&D group of Extreme Vision Ltd, Shenzhen, China.

\bibliographystyle{aaai}
\bibliography{ref}

\begin{thebibliography}{}

\bibitem[\protect\citeauthoryear{Bengio and LeCun}{2016}]{DBLP:conf/iclr/2016}
Bengio, Y., and LeCun, Y., eds.
\newblock 2016.
\newblock {\em The 4th International Conference on Learning Representations,
  {ICLR} 2016, San Juan, Puerto Rico, May 2-4, 2016, Conference Track
  Proceedings}.

\bibitem[\protect\citeauthoryear{Bonawitz \bgroup et al\mbox.\egroup
  }{2019}]{DBLP:journals/corr/abs-1902-01046}
Bonawitz, K.; Eichner, H.; Grieskamp, W.; Huba, D.; Ingerman, A.; Ivanov, V.;
  Kiddon, C.; Konecn{\'{y}}, J.; Mazzocchi, S.; McMahan, H.~B.; Overveldt,
  T.~V.; Petrou, D.; Ramage, D.; and Roselander, J.
\newblock 2019.
\newblock Towards federated learning at scale: System design.
\newblock In {\em CoRR},  http://arxiv.org/abs/1902.01046.

\bibitem[\protect\citeauthoryear{Cheng \bgroup et al\mbox.\egroup
  }{2017}]{DBLP:journals/corr/abs-1710-09282}
Cheng, Y.; Wang, D.; Zhou, P.; and Zhang, T.
\newblock 2017.
\newblock A survey of model compression and acceleration for deep neural
  networks.
\newblock In {\em CoRR},  http://arxiv.org/abs/1710.09282.

\bibitem[\protect\citeauthoryear{Cong \bgroup et al\mbox.\egroup
  }{2019}]{Cong-et-al:2019}
Cong, M.; Weng, X.; Yu, H.; and Qu, Z.
\newblock 2019.
\newblock {FML} incentive mechanism design: Concepts, basic settings, and
  taxonomy.
\newblock In {\em the 1st Interntaional Workshop on Federated Machine Learning
  for User Privacy and Data Confidentiality (FL-IJCAI'19)}.

\bibitem[\protect\citeauthoryear{Doan, Ramakrishnan, and
  Halevy}{2011}]{Doan-et-al:2011}
Doan, A.; Ramakrishnan, R.; and Halevy, A.~Y.
\newblock 2011.
\newblock Crowdsourcing systems on the world-wide web.
\newblock {\em Communications of the ACM} 54(4):86--96.

\bibitem[\protect\citeauthoryear{Dwork}{2006}]{Dwork:2006:DP:2097282.2097284}
Dwork, C.
\newblock 2006.
\newblock Differential privacy.
\newblock In {\em Proceedings of the 33rd International Conference on Automata,
  Languages and Programming (ICALP'06)},  1--12.

\bibitem[\protect\citeauthoryear{Gao \bgroup et al\mbox.\egroup
  }{2019}]{Gao-et-al:2019}
Gao, D.; Liu, Y.; Huang, A.; Ju, C.; Yu, H.; and Yang, Q.
\newblock 2019.
\newblock Privacy-preserving heterogeneous federated transfer learning.
\newblock In {\em Proceedings of the 2019 IEEE International Conference on Big
  Data (IEEE BigData'19)}.

\bibitem[\protect\citeauthoryear{Girshick \bgroup et al\mbox.\egroup
  }{2014}]{RCNN2014}
Girshick, R.; Donahue, J.; Darrell, T.; and Malik, J.
\newblock 2014.
\newblock Rich feature hierarchies for accurate object detection and semantic
  segmentation.
\newblock In {\em Proceedings of the 2014 IEEE Conference on Computer Vision
  and Pattern Recognition (CVPR'14)},  580--587.

\bibitem[\protect\citeauthoryear{Konecn{\'{y}} \bgroup et al\mbox.\egroup
  }{2016}]{45648}
Konecn{\'{y}}, J.; McMahan, H.~B.; Yu, F.~X.; Richtarik, P.; Suresh, A.~T.; and
  Bacon, D.
\newblock 2016.
\newblock Federated learning: Strategies for improving communication
  efficiency.
\newblock In {\em NIPS Workshop on Private Multi-Party Machine Learning}.

\bibitem[\protect\citeauthoryear{Liu \bgroup et al\mbox.\egroup
  }{2019}]{Liu-et-al:2019}
Liu, Y.; Kang, Y.; Yu, H.; Chen, T.; and Yang, Q.
\newblock 2019.
\newblock Secure federated transfer learning.
\newblock In {\em the 1st Interntaional Workshop on Federated Machine Learning
  for User Privacy and Data Confidentiality (FL-IJCAI'19)}.

\bibitem[\protect\citeauthoryear{McMahan \bgroup et al\mbox.\egroup
  }{2016}]{FedAvg2016}
McMahan, H.~B.; Moore, E.; Ramage, D.; and y~Arcas, B.~A.
\newblock 2016.
\newblock Federated learning of deep networks using model averaging.
\newblock In {\em CoRR},  http://arxiv.org/abs/1602.05629.

\bibitem[\protect\citeauthoryear{Redmon and
  Farhadi}{2018}]{DBLP:journals/corr/abs-1804-02767}
Redmon, J., and Farhadi, A.
\newblock 2018.
\newblock Yolov3: An incremental improvement.
\newblock In {\em CoRR},  http://arxiv.org/abs/1804.02767.

\bibitem[\protect\citeauthoryear{Ren \bgroup et al\mbox.\egroup
  }{2017}]{Ren:2017:FRT:3101720.3101780}
Ren, S.; He, K.; Girshick, R.; and Sun, J.
\newblock 2017.
\newblock Faster r-cnn: Towards real-time object detection with region proposal
  networks.
\newblock {\em IEEE Transactions on Pattern Analysis and Machine Intelligence
  (TPAMI)} 39(6):1137--1149.

\bibitem[\protect\citeauthoryear{Voigt and
  Bussche}{2017}]{Voigt:2017:EGD:3152676}
Voigt, P., and Bussche, A. v.~d.
\newblock 2017.
\newblock {\em The EU General Data Protection Regulation (GDPR): A Practical
  Guide}.
\newblock Springer Publishing Company, Incorporated, 1st edition.

\bibitem[\protect\citeauthoryear{Wang \bgroup et al\mbox.\egroup
  }{2019}]{Wang-et-al:2019}
Wang, T.; Zhao, J.; Yu, H.; Liu, J.; Yang, X.; Ren, X.; and Shi, S.
\newblock 2019.
\newblock Privacy-preserving crowd-guided {AI} decision-making in ethical
  dilemmas.
\newblock In {\em Proceedings of the 28th ACM International Conference on
  Information and Knowledge Management (CIKM'19)}.

\bibitem[\protect\citeauthoryear{Wei \bgroup et al\mbox.\egroup
  }{2019}]{Wei-et-al:2019}
Wei, X.; Li, Q.; Liu, Y.; Yu, H.; Chen, T.; and Yang, Q.
\newblock 2019.
\newblock Multi-agent visualization for explaining federated learning.
\newblock In {\em Proceedings of the 28th International Joint Conference on
  Artificial Intelligence (IJCAI'19)},  6572--6574.

\bibitem[\protect\citeauthoryear{Yang \bgroup et al\mbox.\egroup
  }{2019}]{DBLP:journals/tist/YangLCT19}
Yang, Q.; Liu, Y.; Chen, T.; and Tong, Y.
\newblock 2019.
\newblock Federated machine learning: Concept and applications.
\newblock {\em ACM Transactions on Intelligent Systems and Technology (TIST)}
  10(2):12:1--12:19.

\bibitem[\protect\citeauthoryear{Yao}{1982}]{Yao:1982:PSC:1398511.1382751}
Yao, A. C.-C.
\newblock 1982.
\newblock Protocols for secure computations.
\newblock In {\em Proceedings of the 23rd Annual Symposium on Foundations of
  Computer Science (SFCS'82)},  160--164.

\bibitem[\protect\citeauthoryear{Yao}{1986}]{Yao:1986:GES:1382439.1382944}
Yao, A. C.-C.
\newblock 1986.
\newblock How to generate and exchange secrets.
\newblock In {\em Proceedings of the 27th Annual Symposium on Foundations of
  Computer Science (SFCS'86)},  162--167.

\bibitem[\protect\citeauthoryear{Yu \bgroup et al\mbox.\egroup
  }{2017}]{Yu-et-al:2017SciRep}
Yu, H.; Miao, C.; Chen, Y.; Fauvel, S.; Li, X.; and Lesser, V.~R.
\newblock 2017.
\newblock Algorithmic management for improving collective productivity in
  crowdsourcing.
\newblock {\em Scientific Reports} 7:doi:10.1038/s41598--017--12757--x.

\bibitem[\protect\citeauthoryear{Yu \bgroup et al\mbox.\egroup
  }{2018}]{Yu-et-al:2018IJCAI}
Yu, H.; Shen, Z.; Miao, C.; Leung, C.; Lesser, V.~R.; and Yang, Q.
\newblock 2018.
\newblock Building ethics into artificial intelligence.
\newblock In {\em Proceedings of the 27th International Joint Conference on
  Artificial Intelligence (IJCAI'18)},  5527--5533.

\bibitem[\protect\citeauthoryear{Zhao \bgroup et al\mbox.\egroup
  }{2018}]{DBLP:journals/corr/abs-1807-05511}
Zhao, Z.; Zheng, P.; Xu, S.; and Wu, X.
\newblock 2018.
\newblock Object detection with deep learning: {A} review.
\newblock In {\em CoRR},  http://arxiv.org/abs/1807.05511.

\end{thebibliography}

\end{document}